\documentclass[letterpaper, 10 pt, conference]{ieeeconf}  

\IEEEoverridecommandlockouts

\usepackage{subfigure}

\usepackage{graphics} 
\usepackage{epsfig} 
\usepackage{amssymb}  

\usepackage{booktabs}
\usepackage{multirow}
\usepackage{pifont}
\usepackage{cite}

\usepackage{balance}

\usepackage{latexsym}
\usepackage{bbm}
\usepackage{graphics}
\usepackage{amsmath, amssymb}
\usepackage{epsfig}

\def\statespace {{\cal S}}
\def\mdp {{\cal M}}
\def\actionspace {{\cal A}}
\def\probs {{\cal P}}
\def\transitionmodel {{\cal T}}

\usepackage{tikz}
\usepackage{textcomp}
\usepackage{hyperref}
\usepackage{lipsum}

\newcommand\copyrighttext{%
  \footnotesize \textcopyright 2020 IEEE. Personal use of this material is permitted. Permission from IEEE must be obtained for all other uses, in any current or future media, including reprinting/republishing this material for advertising or promotional purposes, creating new collective works, for resale or redistribution to servers or lists, or reuse of any copyrighted component of this work in other works.} 
\newcommand\copyrightnotice{%
\begin{tikzpicture}[remember picture,overlay]
\node[anchor=south,yshift=10pt] at (current page.south) {\fbox{\parbox{\dimexpr\textwidth-\fboxsep-\fboxrule\relax}{\copyrighttext}}};
\end{tikzpicture}%
}

\title{\LARGE {\bf CityLearn: Diverse Real-World Environments for Sample-Efficient Navigation Policy Learning}}

\author{Marvin Chanc\'an$^{1,2}$ and Michael Milford$^{1}$
\thanks{The work of M.C. was supported by the Peruvian Government. The work of M.M. was supported by ARC grants FT140101229, CE140100016, the QUT Centre for Robotics, and the Australian Government via grant AUSMURIB000001 associated with ONR MURI grant N00014-19-1-2571.}
\thanks{$^{1}$ QUT Centre for Robotics, School of Electrical Engineering and Robotics, Queensland University of Technology, Brisbane, Australia}
\thanks{$^{2}$ School of Mechatronics Engineering, Universidad Nacional de Ingenier\'ia, Lima, Peru. {\tt\small mchancanl@uni.pe}}}

\begin{document}

\maketitle
\copyrightnotice
\thispagestyle{empty}
\pagestyle{empty}

\begin{abstract}
Visual navigation tasks in real-world environments often require both self-motion and place recognition feedback. While deep reinforcement learning has shown success in solving these perception and decision-making problems in an end-to-end manner, these algorithms require large amounts of experience to learn navigation policies from high-dimensional data, which is generally impractical for real robots due to sample complexity. In this paper, we address these problems with two main contributions. We first leverage place recognition and deep learning techniques combined with goal destination feedback to generate compact, bimodal image representations that can then be used to effectively learn control policies from a small amount of experience. Second, we present an interactive framework, \textit{CityLearn}, that enables for the first time training and deployment of navigation algorithms across city-sized, realistic environments with extreme visual appearance changes. CityLearn features more than 10 benchmark datasets, often used in visual place recognition and autonomous driving research, including over 100 recorded traversals across 60 cities around the world. We evaluate our approach on two CityLearn environments, training our navigation policy on a single traversal. Results show our method can be over 2 orders of magnitude faster than when using raw images, and can also generalize across extreme visual changes including day to night and summer to winter transitions.
\end{abstract}

\section{Introduction}

The ability to sense location in time and space is key, for both robots and living beings, to enable navigation in highly dynamic real-world environments. For mobile robots, the way they can create a particular, internal world representation often depends on their perceptual limitations as well as how they interact and make decisions with the environment \cite{IEEEexample:st-book}. Visual feedback provides high-dimensional information that, when encoded properly, can be used to make sense of where they are and where they need to go. Similarly, self-motion feedback also provides information concerning the current position within an environment. These two sensory input modalities are concurrent, time-aligned and often complementary during goal-driven navigation tasks.

Recent deep reinforcement learning (RL) approaches have successfully performed active navigation tasks on simulated environments using real-world street imagery \cite{IEEEexample:nomap} or synthetic scenarios \cite{IEEEexample:complex,IEEEexample:banino2018vector,IEEEexample:cueva2018emergence}. These algorithms, however, generally utilize additional feedback data, e.g. the agent-relative velocity or reward function values, that eventually increase their network policy architecture and sample complexity.

\begin{figure}[!t]
   \centering
   \includegraphics[width=\columnwidth]{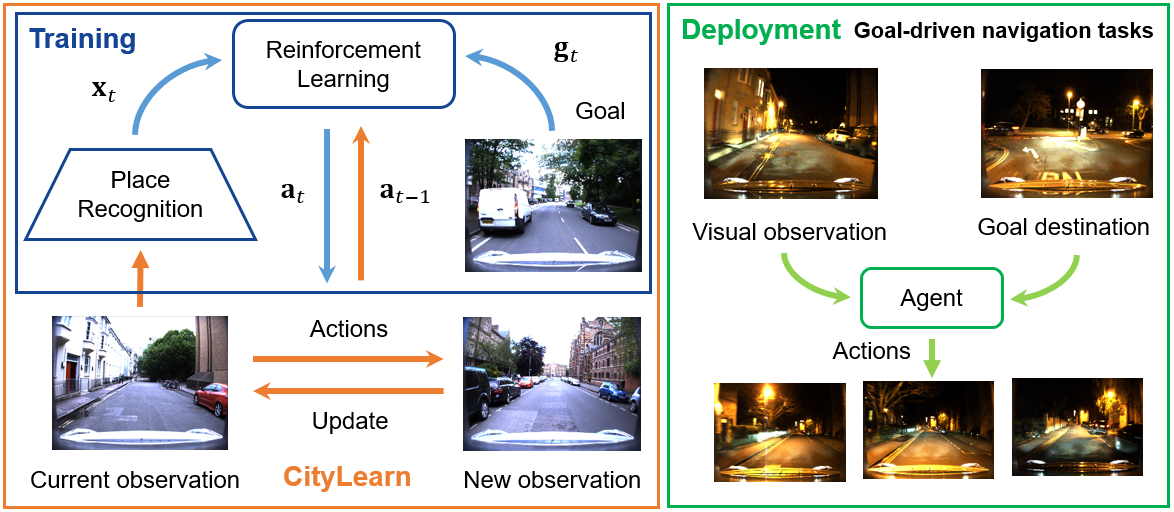}
   \vspace{-7mm}
   \caption{\textbf{The CityLearn framework}. We leverage VPR and RL methods to learn control policies for goal-driven navigation tasks. Our method is efficient and can generalize across extreme environmental changes.}
   \label{approach}
   \vspace{-2mm}
\end{figure}

\begin{figure}[!t]
\centering
\includegraphics[width=0.95\columnwidth]{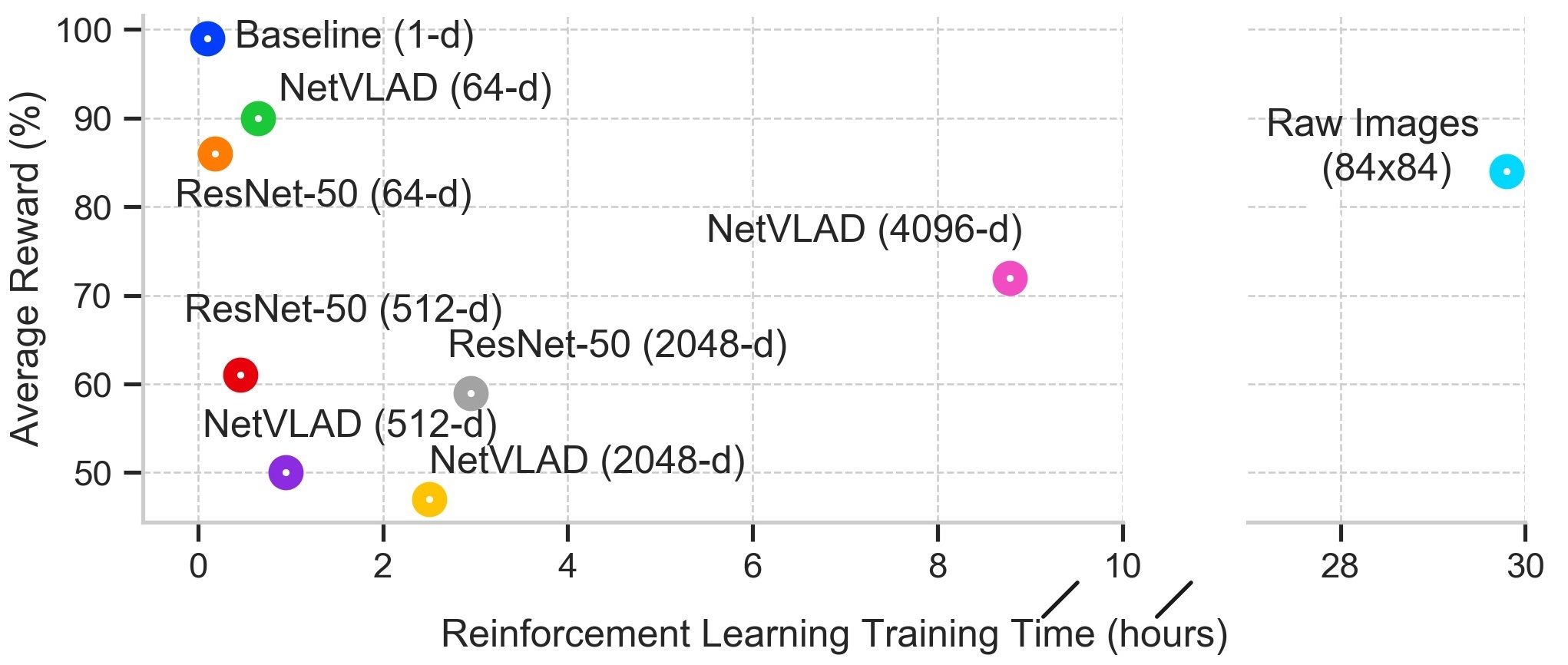}
\vspace{-3mm}
\caption{\textbf{Performance and compute characterization}. RL training results for (i) our approach using off-the-shelf VPR (NetVLAD) and deep learning (ResNet-50) models with a number of feature dimensions (e.g. 64, 512, 2048, 4096), (ii) a baseline agent that uses 1-d position feedback instead of images, and (iii) an agent trained end-to-end using raw images.}
\label{rl-tradeoff}
\vspace{-6mm}
\end{figure}

Visual place recognition (VPR) models, on the other hand, are required to successfully match two image sequences of recorded data in real-world environments. While recent improvements using deep learning \cite{IEEEexample:nvlad, IEEEexample:zetao2017, IEEEexample:lost} and algorithmic methods \cite{IEEEexample:mpf} have contributed to state-of-the-art results on city-sized datasets, whether those models can enable navigation capabilities on real robots is not well explored.

In this work, we leverage both VPR and RL techniques to efficiently learn control policies for navigation tasks (Fig. \ref{approach}). Our resulting control policy is able to perform goal-driven navigation tasks using only two sensory feedback modalities (goal destination and visual representations). The results demonstrate that our policy is able to generalize over a range of extreme environmental changes on real-world datasets, while drastically reducing the amount of training experience, e.g. from 29h48m to 11m (Fig. \ref{rl-tradeoff}). We also show how our approach can achieve practical sample efficiency in an interactive and diverse environment that we call CityLearn.\footnote{Project page: {\tt mchancan.github.io/projects/CityLearn}}

The main contributions of this paper are:
\begin{enumerate}
\item CityLearn: An interactive open framework with real-world environments for perception and decision-making to enable the evaluation of navigation algorithms on more than 10 robotic benchmark datasets with challenging environmental transitions (Fig. \ref{citylearn-envs}).
\item A new approach to sample-efficient RL training for goal-driven navigation tasks. We use VPR and deep learning models to encode our sensory input images which, when combined with goal destination signals, can generate compact, bimodal representations, from which a navigation policy can be learned to generalize across extreme visual changes such as day to night or summer to winter cycles.
\end{enumerate}

\section{Related Work}

In robotics research, the use of probabilistic techniques played an important role in solving robotic problems such as how the robot's sensory information should be integrated to generate internal states and support the decision-making process \cite{IEEEexample:st-book}. In the mid-1990s, these methods allowed the deployment of navigation algorithms on real robots by using conditional probability distributions, instead of deterministic functions at a fixed time interval (as in classical control), to compute more general control actions that govern the robot's states \cite{IEEEexample:nav101, IEEEexample:nav102, IEEEexample:st98}. In the same decade, moreover, the field of RL started to attract the interest of roboticists. RL agents learn specific behavior through interactions with dynamic environments only by reward and punishment signals \cite{IEEEexample:lin}. Therefore, the use of RL to solve more complex robot-learning problems started to be extended with the incorporation of neural networks to obtain broader generalization capabilities. Ideas like hierarchical or curriculum learning \cite{IEEEexample:c7} were also proposed to reduce the learning time and solve these complex, physically realistic robotic problems in simulation environments.

\subsection{Deep Reinforcement Learning based Navigation}

The spread of \textit{convolutional neural networks} (CNN) has yielded impressive state-of-the-art results in computer vision, natural language processing, and many other related domains over the past eight years \cite{IEEEexample:cnn1}. Similarly, recent research incorporating deep neural networks to more advanced RL algorithms for navigation tasks have shown promising results in simulated environments.

Recent works have trained deep RL agents or deep-learning-based models to perform navigation tasks using real-world images \cite{IEEEexample:nomap, IEEEexample:deepnav, IEEEexample:c20, IEEEexample:carballo2018end, IEEEexample:amini2019variational, IEEEexample:xu2017end, IEEEexample:c12, IEEEexample:c13, IEEEexample:c14, IEEEexample:dhiman2018critical}; typically generalizing well over different visual conditions with minimal additional training and network architecture changes. Similarly, researchers have used non end-to-end RL approaches with real data that, when encoded via off-the-shelf deep learning models, can efficiently learn navigation policies in unstructured environments \cite{IEEEexample:deplo}. Related deep RL approaches have also shown success on target-driven navigation tasks but have only been demonstrated in indoor \cite{IEEEexample:c16, IEEEexample:gupta2017cognitive, IEEEexample:Kolve2017AI2THORAI, IEEEexample:savva2017minos,  IEEEexample:kahn2018composable, IEEEexample:brunner2018teaching, IEEEexample:wu2018building, IEEEexample:bansal2019combining, IEEEexample:wang2018omnidirectional, IEEEexample:mousavian2019visual} or synthetic environments \cite{IEEEexample:complex,IEEEexample:banino2018vector,IEEEexample:cueva2018emergence, IEEEexample:c31}. 

These approaches, however, often use additional feedback data such as reward function values or the agent-relative velocity that further increase the policy network size and training requirements. These factors also increase the number of interactions required with the environment, typically to the order of millions of episodes. These systems, moreover, are often evaluated on the same environment used for training, thus their generalization capabilities to different visual conditions are often unknown; alternatively, it is necessary to increase the complexity of their architectures for them to successfully train and generalize to challenging environmental conditions.



\subsection{Visual Place Recognition}\label{vpr}

\begin{figure}[!t]
\includegraphics[width=\columnwidth]{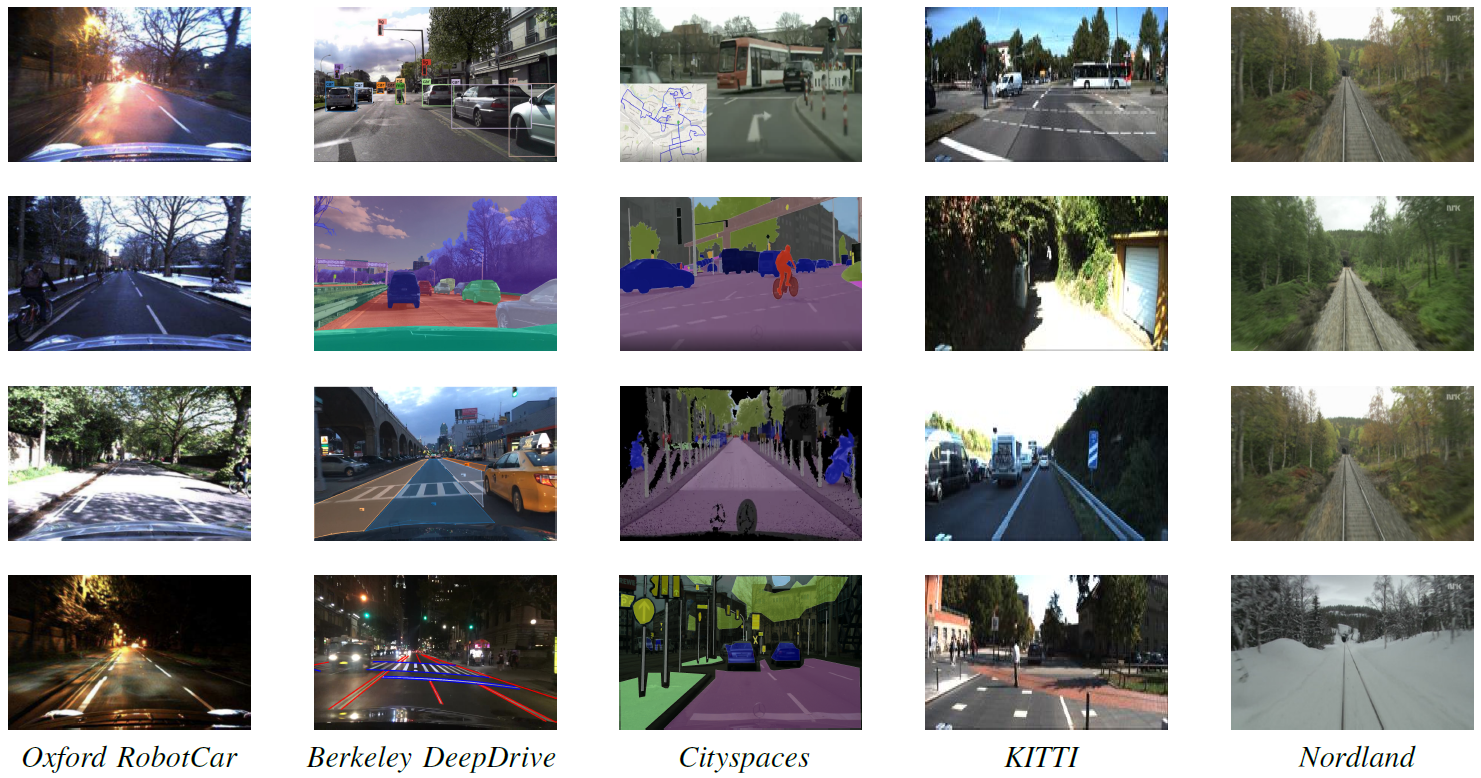}
\vspace{-6mm}
\caption{Five selected benchmark datasets that can be used in CityLearn.}
\label{citylearn-envs}
\vspace{-4mm}
\end{figure}

VPR methods for sequence-based localization tasks typically perform a multi-frame matching procedure between two or more traversals (query and reference) on stationary real-world datasets. Both query and reference sequence of images often include challenging appearance and viewpoint changes between them (e.g. different weather or seasonal conditions), illumination changes due to time of day, and dynamic objects (Fig. \ref{citylearn-envs}). A VPR algorithm for sequence-based datasets can be broadly split into two main steps \cite{IEEEexample:seqslam, IEEEexample:chen2014convolutional, IEEEexample:naseer2014robust, IEEEexample:pepperell2014all, IEEEexample:milford2014condition, IEEEexample:vprsurvey, IEEEexample:sunderhauf2015performance, IEEEexample:cnnlanmark, IEEEexample:ferrarini2019visual, IEEEexample:lost, IEEEexample:merrill2018lightweight, IEEEexample:lowry2018lightweight, IEEEexample:sattler2018benchmarking, IEEEexample:mount2019automatic, IEEEexample:arelarge, IEEEexample:hausler2019filter, IEEEexample:lowry2019similarity, IEEEexample:khaliq2019holistic, IEEEexample:mpf, IEEEexample:chancan2020hybrid}: (1) feature extraction process utilizing either hand-crafted or deep-learning-based techniques to obtain compact image representations, that can then be (2) matched via conventional similarity metrics (e.g. cosine or $L_2$ distance) or more elaborate multi-frame temporal filtering algorithms such as SeqSLAM \cite{IEEEexample:seqslam} and many others \cite{IEEEexample:pepperell2014all, IEEEexample:openseqslam2, IEEEexample:garg2018don}. Though recent improvements using temporal filtering approaches have shown state-of-the-art results \cite{IEEEexample:mpf}, we note that those methods need to have at least two traversals from the same environment at the time of performing their final matching procedure.

In this work, for the proposed goal-oriented navigation task, we use a single traversal to train our control policy network, which is then evaluated on the remaining traversals. Consequently, we choose VPR and deep learning techniques that are known to obtain better, compact visual representations from raw images, such as ResNet \cite{IEEEexample:resnet} or NetVLAD \cite{IEEEexample:nvlad}---which performs well compared to related VPR models \cite{IEEEexample:levelling}---rather than full VPR models that use algorithmic techniques on top of those deep learned representations.

\begin{table*}[!t]
\caption{CityLearn: Detailed comparison with recent real-world environments}
\vspace{-4mm}
\label{comparison}
\begin{center}
\begin{tabular}{ccccccccc}
\toprule
Environment & region/dataset & \#trav & \#imgs & av. step & \#sensors & journey & city & country \\
\midrule
\multirow{6}{*}{StreetLearn} & Wall Street & 1 & \multirow{2}{*}{56k} & 9.8m & \multirow{5}{*}{1$\times$panoramic cam.} & \multirow{2}{*}{548.8km} & \multirow{2}{*}{New York} & \multirow{6}{*}{USA} \\
\multirow{6}{*}{\cite{IEEEexample:mirowski2019streetlearn}}& Union Square & 1 & & 9.8m & \multirow{5}{*}{(360$^{\circ}$view)}  & \\
\cmidrule{2-5} \cmidrule{7-8}
& Hudson Rive & 1 & \multirow{4}{*}{58k} & 9.9m & & \\
& CMU & 1 & & 9.9m & & \multirow{2}{*}{574.2km} & \multirow{2}{*}{Pittsburgh} \\
& Allergheny & 1 & & 9.8m & & \\
& South Shore & 1 & & 9.9m &  & \\
\midrule
\multirow{12}{*}{\textbf{CityLearn}} & \multirow{2}{*}{Oxford RobotCar} & \multirow{2}{*}{133} & \multirow{2}{*}{20M} & \multirow{2}{*}{0.2m} & \multirow{1}{*}{6$\times$stereo cam. (360$^{\circ}$view)} & \multirow{2}{*}{10km} & \multirow{2}{*}{Oxford} & \multirow{2}{*}{UK}\\
\multirow{12}{*}{[Ours]} &&&& & \multirow{1}{*}{1$\times$3D and 2$\times$2D LiDAR} && \\
\cmidrule{2-9}
& Berkeley DeepDrive & 100k & 120M & 30fps & 1$\times$cam. (front) & 1100h & multiple & USA \\
\cmidrule{2-9}
& Cityspaces & 50 & - & - & 1$\times$cam. (front) & 100h & multiple (50) & \multirow{2}{*}{Germany} \\
\cmidrule{2-8}
& KITTI & 22 & - & 10fps & 4$\times$cam. (front, rear) & 6h & Karlsruhe & \\
\cmidrule{2-9}
& Nordland Railway & 4 & 3.6M & 0.05m & 1$\times$cam. (front) & 728km & Trondheim--Bod\o & Norway \\
\cmidrule{2-9}
& \multirow{1}{*}{Multi-Lane Road} & 4 & - & - & 1$\times$cam. (front) & 4km & Gold Coast (GC) & \multirow{4}{*}{Australia} \\
& \multirow{1}{*}{Gold Coast Drive$^\dagger$} & 1 & - & - & 1$\times$cam. (front) & 87km & Brisbane--GC & \\
& \multirow{1}{*}{UQ St. Lucia} & 1 & - & - & 1$\times$cam. (front) & 9.5km & Brisbane & \\
& \multirow{1}{*}{St. Lucia Multiple Times} & 10 & - & 15fps & 1$\times$cam. (front) & - & Brisbane & \\
& \multirow{1}{*}{Alderley Day/Night$^\ddagger$} & 2 & 31.5k & - & 1$\times$cam. (front) & 16km & Brisbane & \\
\bottomrule
\end{tabular}
\raggedright
\scriptsize{$^\dagger$Provides frame correspondences or $^\ddagger$reference trajectory instead of GPS data.}
\end{center}
\vspace{-4mm}
\end{table*}

\section{The CityLearn environment}

VPR methods are often evaluated on variety-rich, real-world datasets collected over long traversals across different seasons, time of day or weather conditions, including dynamic objects, such as cars, traffic, and pedestrians, along with longer-term changes such as construction or roadworks \cite{IEEEexample:seqslam, IEEEexample:oxford, IEEEexample:lookonce, IEEEexample:cnnlanmark, IEEEexample:fabrat, IEEEexample:seqsearch, IEEEexample:kitti, IEEEexample:naseer, IEEEexample:guo2018safe, IEEEexample:caesar2019nuscenes} (Fig. \ref{citylearn-envs}). The data obtained typically includes videos or sequences of images providing panoramic or 360$^{\circ}$ views from stereo cameras, scans of 2D/3D Lidar sensors, visual odometry data, and GPS/inertial data that can then be used as ground truth labels.

We leverage those real-world datasets to create CityLearn, an interactive open framework that enables, for the first time, the training and testing of navigation algorithms on city-sized, realistic environments. Our fully-configurable environment runs on top of the Unity game engine and their ML-Agents framework \cite{IEEEexample:juliani2018unity}. CityLearn is related to the recent StreetLearn work \cite{IEEEexample:mirowski2019streetlearn} used in \cite{IEEEexample:nomap,IEEEexample:hermann2019learning, IEEEexample:li2019cross} but has a range of useful differences. We propose the usage of diverse environments across $5$ countries and additionally enable loading any other dataset including in-house recorded data; see Tables \ref{comparison} and \ref{support} for a detailed comparison. 

\begin{table}[!t]
\caption{StreetLearn vs. CityLearn: Support and Features}
\vspace{-4mm}
\label{support}
\begin{center}
\begin{tabular}{lcc}
\midrule
Description & StreetLearn \cite{IEEEexample:mirowski2019streetlearn} & \textbf{CityLearn} [Ours]\\
\midrule
Operating system & Ubuntu 18.04 & Windows/Linux/Mac \\
Environment engine & StreetLearn & Unity/ML-Agents\\
Language/ML frameworks & C++, Python/TF & C\#, Python/TF\\
Min. RAM per env. & 12GB & 2GB\\
Number of public datasets & 1 & 10+ \\
Number of cities & 2 & 60+\\
Number of traversals & 1 & 100+\\
Min. average agent step & 9.8m & 0.05m\\
Multi-environment training & \ding{52} & \ding{52}\\
Feature public datasets & \ding{56} & \ding{52}\\
Appearance changes & \ding{56} & \ding{52}\\
Viewpoint changes & \ding{52} & \ding{52}\\
Multiple times of day & \ding{56} & \ding{52}\\
Multiple weather/seasons & \ding{56} & \ding{52}\\
\midrule
\end{tabular}
\end{center}
\vspace{-6mm}
\end{table}

In Table \ref{comparison}, each environment (region/dataset) also includes GPS data; except for Goald Coast Drive and Alderley Day/Nigth. Related frameworks for city-scale navigation based on real-world images were not considered in Table \ref{comparison} as they interact differently with the environment via natural language communication \cite{IEEEexample:c12,IEEEexample:c13,IEEEexample:c14}. 

\section{Problem Statement and Methods}

Our goal is to train a policy network to perform goal-driven navigation tasks. To enable sample-efficiency, we use either off-the-shelf VPR or deep learning models to encode our sensory input images and obtain multi-dimensional feature vectors. Then, using RL, we combine these features with compact goal destinations, resulting in compact, bimodal representations that can then be used to train our policy using a single traversal in our CityLearn environment.

We use a Markov Decision Process $\mdp$ with discrete state $\mathbf{s}_t \in \statespace$ and action $\mathbf{a}_t \in \actionspace$ spaces, and a transition operator $\transitionmodel: \statespace \times \actionspace \to \statespace$ to model our navigation tasks as a finite-horizon $T$ problem.
Our goal is to find $\theta^*$ that maximizes the objective function: 

\begin{equation}
\label{objective}
J(\theta) = \mathbb{E}_{\tau \sim \pi_\theta(\tau)} \left[\sum_{t=1}^T \gamma r(\tau)\right]
\end{equation}

where $\pi_\theta: \statespace \to \probs(\actionspace)$ is the stochastic navigation policy we want to learn, and $r: \statespace \times \actionspace \to \mathbb{R}$ is the reward function with discount factor $\gamma$. To optimize $\pi_\theta$, we parameterize it with a neural network that learns $\theta$, as described in  Sec. \ref{policy-learn}. $\statespace$ is defined by our compact, bimodal space representation $\mathbf{b}_t$  generated by combining the agent's visual observation $\mathbf{x}_t$ and a 1-\textit{d} goal destination $\mathbf{g}_t$, as also detailed in Secs. \ref{feat-obs} and \ref{policy-learn}. $\actionspace$ is defined over discrete action movements in the agent's action space $\mathbf{a}_t$. We evaluate our approach on two challenging CityLearn environments with extreme visual changes such as day to night for Oxford RobotCar \cite{IEEEexample:oxford}, and summer to winter for the Nordland \cite{IEEEexample:nordland} dataset (Fig. \ref{datasets}).

\subsection{Visual Observations}\label{feat-obs}

We encode our sensory input images -- which are either 1920$\times$1080 RGB for the Nordland dataset or 1280$\times$960 RGB for the Oxford RobotCar dataset -- using either off-the-shelf VPR (NetVLAD) or deep learning (ResNet-50) models. For NetVLAD \cite{IEEEexample:nvlad}, we use their best performing network, based on VGG-16 \cite{IEEEexample:simonyan2014very} with PCA plus whitening, to encode our images into a range of visual observations consisting of 4096-\textit{d}, 2048-\textit{d}, 512-\textit{d} and 64-\textit{d} feature vectors. For ResNet-50 \cite{IEEEexample:resnet}, we use a network trained on ImageNet \cite{IEEEexample:imagenet} to extract image representations of 2048-\textit{d}, which we then reduce to more compact representations such as 512-\textit{d} and 64-\textit{d} using the algorithm provided in NetVLAD for dimensionality reduction.

Once we obtain our visual observations, $\mathbf{x}_t$, we combine them with a 1-\textit{d} goal destination, $\mathbf{g}_t$, to generate our compact, bimodal representation, $\mathbf{b}_t$, that serves as input to our navigation policy, see Fig. \ref{policy}(b). $\mathbf{g}_t$ is encoded as a 1-\textit{d} feature vector to preserve the compactness of our final bimodal representation $\mathbf{b}_t$ which are feature vectors of 65-\textit{d}, 513-\textit{d}, 2049-\textit{d}, and 4097-\textit{d}.

\subsection{Policy Learning for Visual Navigation}\label{policy-learn}

Our objective is to learn a policy for goal-directed navigation tasks using a compact, bimodal representation such as $\mathbf{b}_t$. While there has been some success using deep reinforcement learning for navigation tasks from raw images \cite{IEEEexample:nomap, IEEEexample:complex}, they require the addition of more feedback modalities (e.g. reward values or agent's velocity) that eventually increase the number of interactions with the environment and training time. We aim to investigate the performance of using $\mathbf{b}_t$, obtained in Sec. \ref{feat-obs}, to train our policy.

\begin{figure}[!t]
   \centering
   \includegraphics[width=\columnwidth]{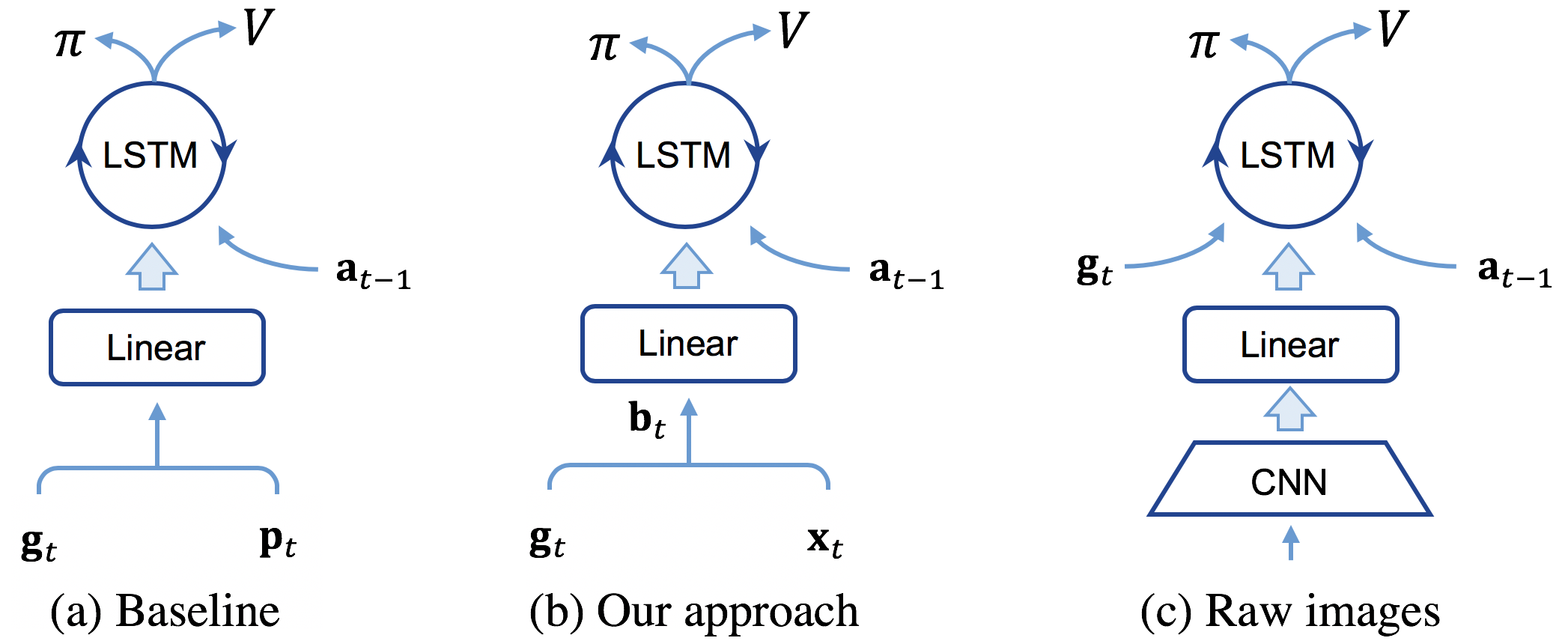}
   \vspace{-6mm}
   \caption{\textbf{Navigation baseline agents}. {\tt Our approach}  (b) uses the goal destination $\mathbf{g}_t$ and compact visual observations $\mathbf{x}_t$ to generate bimodal representations $\mathbf{b}_t$ which can then be combined with the agent's previous action $\mathbf{a}_{t-1}$ to estimate a stochastic navigation policy $\pi$ and value function $V$. We also train a {\tt baseline} (a) agent using its current position $\mathbf{p}_t$ instead of $\mathbf{x}_t$, and another agent using {\tt raw images} (c) from scratch.}
   \label{policy}
   \vspace{-4mm}
\end{figure}

\textbf{Task setup}: We design a navigation task where a successful task requires reasoning using our visual observations and goal destination $\mathbf{b}_t$ to find a required target $\mathbf{g}_t$ over a single traversal in the CityLearn environment (Figs. \ref{approach} and \ref{policy}). 

\textbf{Our approach}: We choose the proximal policy optimization (PPO) algorithm \cite{IEEEexample:ppo} to optimize our objective function in Eq. \eqref{objective}. PPO is a variation of TRPO \cite{IEEEexample:trpo} that constraints the policy update, while striking the balance between sample complexity and hyperparameter tuning to achieve state-of-the-art results on a range of benchmark RL problems. {\tt Our agent} network architecture, see Fig. \ref{policy}(b), comprises of a single linear \textit{multi-layer perceptron} (MLP) of 512 units that encodes $\mathbf{b}_t$, see Fig. \ref{policy}(b). We then combine it with the agent's previous action, $\mathbf{a}_{t-1}$, using a single recurrent layer long short-term memory (LSTM) \cite{IEEEexample:lstm} of 256 units, to estimate the required actions from the estimated policy $\pi$ and the value function $V$. Additionally, we implement two policy networks for comparison purposes, also shown in Figs. \ref{policy}(a) and (c), whose details are provided in Sec. \ref{baselines}.

\textbf{Reward design and curriculum learning}: We use $7$ levels of curriculum learning \cite{IEEEexample:c7} to encourage the agent to explore the environment gradually in order to find increasingly distant destinations \cite{IEEEexample:nomap}. Our sparse reward function gives the agent a reward of $+1$ only when it finds the target, potentially receiving a punishment of $-1/{\tt ms}$ when it heads away from the required destination, with ${\tt ms}$ being the maximum number of agent steps per episode.

\subsection{Baseline Agents}\label{baselines}

We compare {\tt our approach}, described in Section \ref{policy-learn}, against two additional agent architectures: {\tt baseline} and {\tt raw images}, as shown in Figs. \ref{policy}(a) and (c), respectively. In all our experiments, the goal destination, $\mathbf{g}_t$, is encoded using a 1-\textit{d} feature vector for fair comparison (Fig. \ref{policy}), but it can easily be adapted to use more complex encoding methods, as per previous work \cite{IEEEexample:nomap}. The code of the three RL baseline agents including {\tt our approach}, shown in Fig. \ref{policy}, is made publicly available along with CityLearn.

\textbf{Baseline}: This {\tt baseline} agent is a relatively trivial baseline, see Fig. \ref{policy}(a), that uses a 1-\textit{d} feature vector as its current position $\mathbf{p}_t$, instead of $\mathbf{x}_t$. While this substantially simplifies the problem, it is a competitive agent reference since it achieves 100\% completed tasks on deployment.

\textbf{Raw images}: Our {\tt raw images} agent uses a CNN visual module of $2$ \textit{convolutional} layers, see Fig. \ref{policy}(c), as in previous works \cite{IEEEexample:impala, IEEEexample:complex}. The first CNN layer has a kernel of size $8\times8$, a stride of $4\times4$, and 16 feature maps. The second CNN layer has a kernel of size $4\times4$, a stride of $2\times2$, and 32 feature maps. The input consisted of RGB images of $84\times84$.


\subsection{Evaluation Metrics}\label{metrics}

\begin{figure*}[!t]
   \centering
   \subfigure{
   \includegraphics[width=\columnwidth]{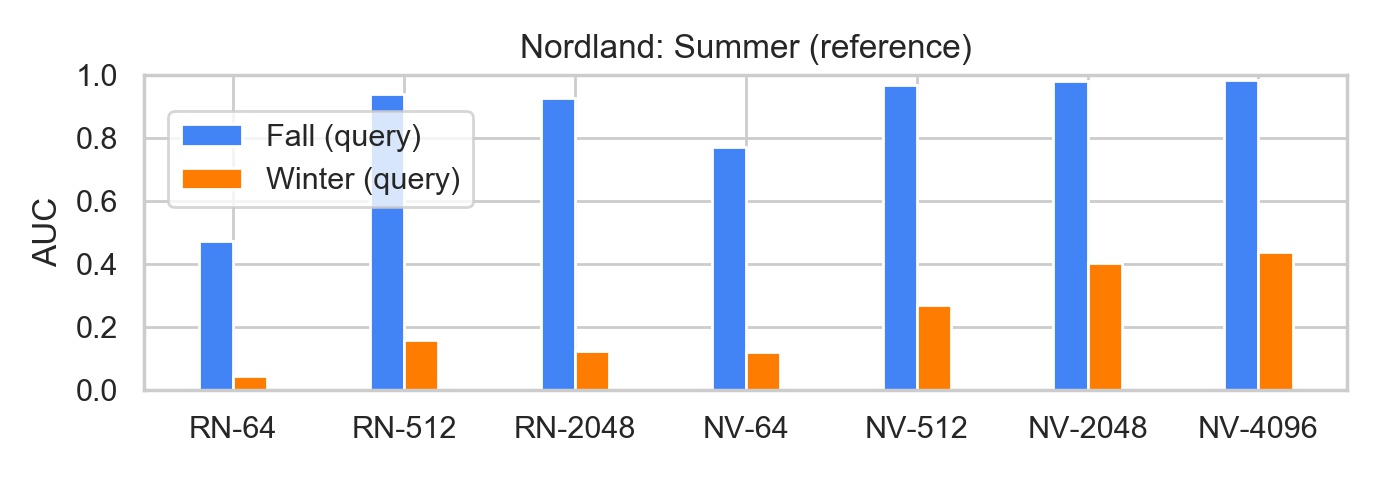}}
   \subfigure{
   \includegraphics[width=\columnwidth]{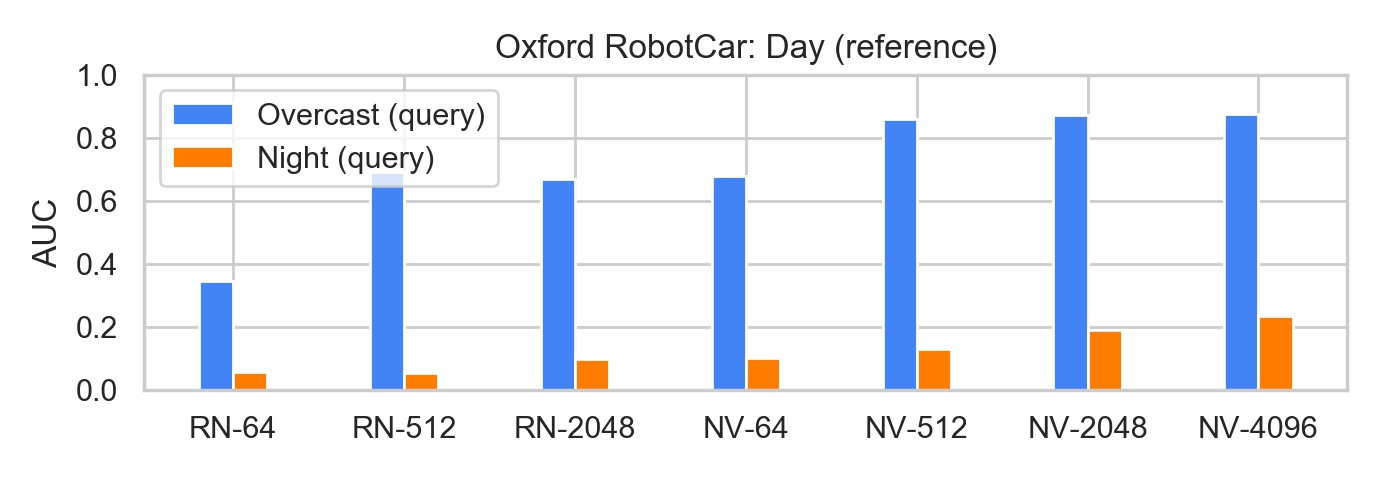}}
   \vspace{-4mm}
   \caption{\textbf{AUC place recognition performance}, on the \textbf{Nordland} (left) and \textbf{Oxford Robotcar} (right) datasets, evaluated under moderate and extreme environmental changes. We use off-the-shelf place recognition (NetVLAD: NV) and deep learning (ResNet-50: RN) models to encode our RGB images into a range of feature dimensions: 64-\textit{d}, 512-\textit{d}, 2048-\textit{d}, 4096-\textit{d}).}
   \label{auc}
\end{figure*}

\textbf{Visual place recognition}: VPR performance using our encoded visual observations are reported via area under the curve (AUC) metrics across a number of feature dimensions (Fig. \ref{auc}). We train a classifier on each reference traversal using a single MLP that receives our encoded visual observations. We then use this trained classifier to evaluate the remaining query traversals. Once we have the scores for both query and reference, we compute the precision-recall curves from where we can obtain the overall AUC performance.

\begin{figure*}[!ht]
\includegraphics[width=\textwidth, height=14mm]{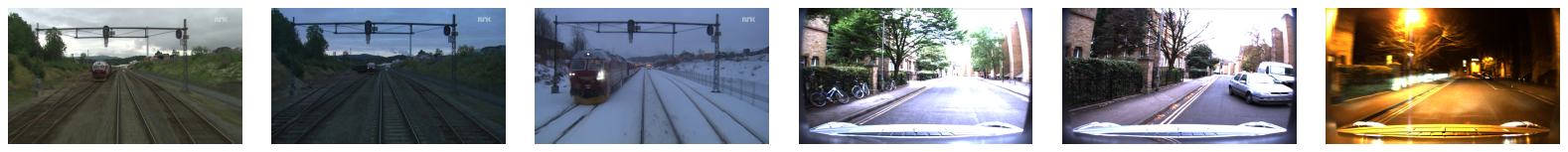}
\vspace{-6mm}
\caption{\textbf{Two diverse real-world benchmark datasets} used in our experiments. The \textbf{Nordland} dataset (left to right-center) including summer, fall, and winter traversals. The \textbf{Oxford RobotCar} dataset (left-center to right) including day, overcast, and night traversals.}
\vspace{-4mm}
\label{datasets}
\end{figure*}

\begin{figure*}[!t]
\centering
\subfigure{\includegraphics[width=6.25in]{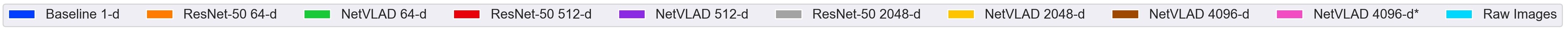}}
\subfigure{\includegraphics[width=3.4in]{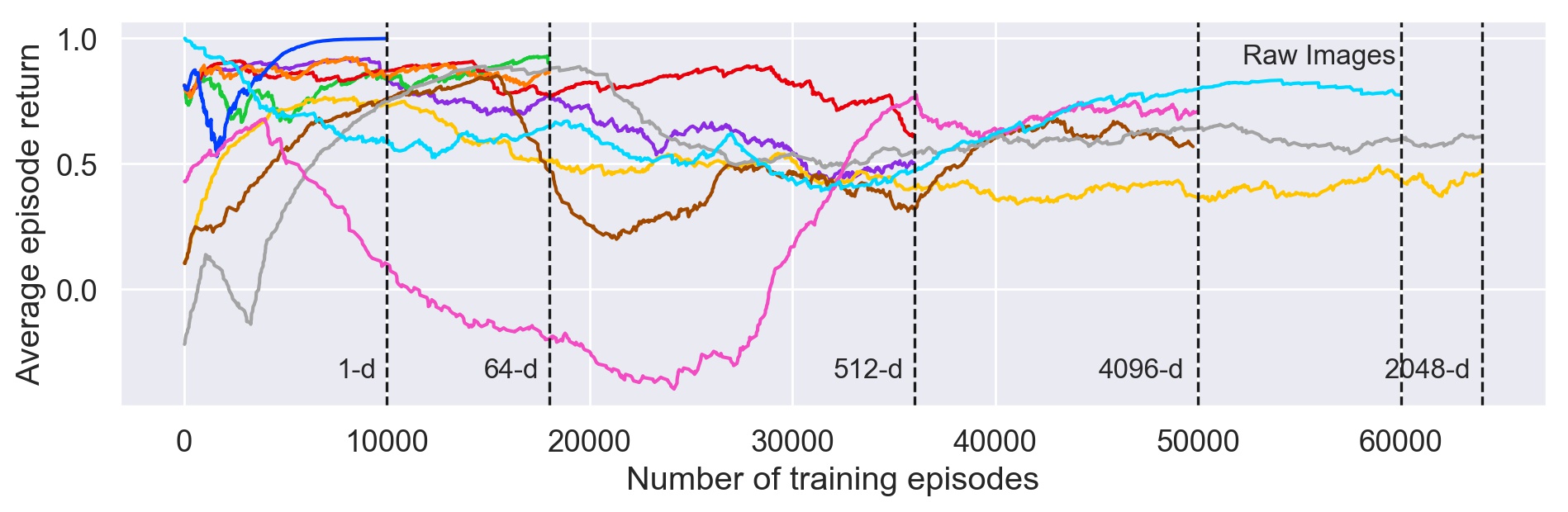}}   
\subfigure{
\includegraphics[width=3.4in]{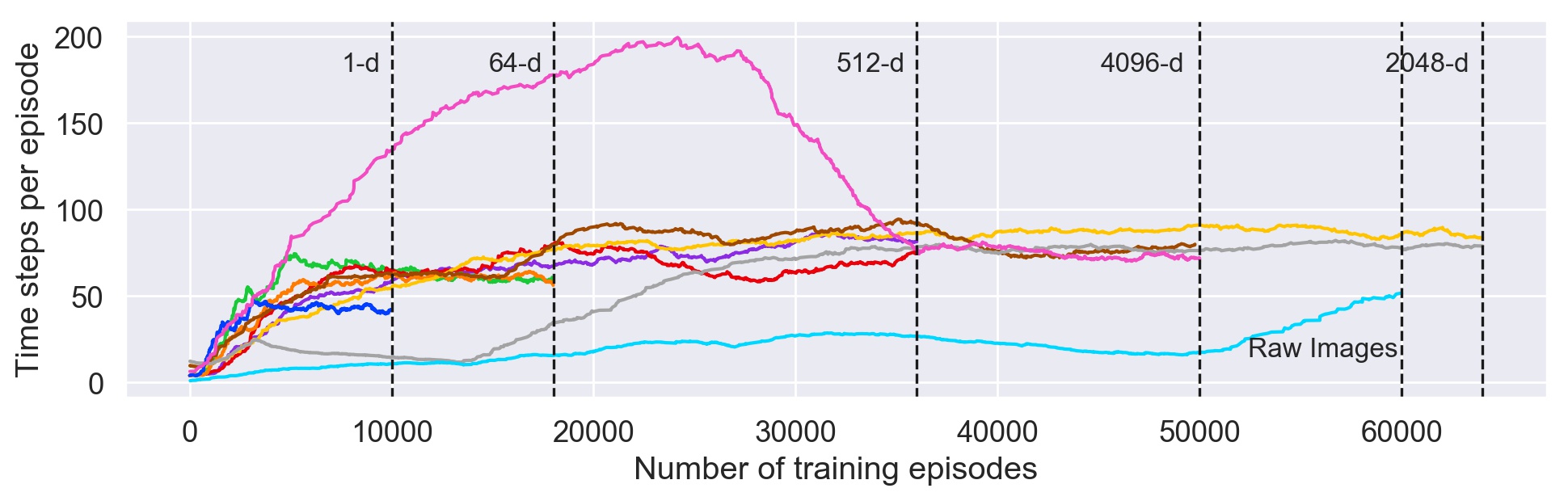}}
\vspace{-2mm}
\caption{\textbf{RL training curves}. {\tt Our approach} uses NetVLAD and ResNet-50 models, with 64-\textit{d}, 512-\textit{d}, 2048-\textit{d}, 4096-\textit{d} feature representations, can efficiently train our RL navigation policy compared to an agent trained end-to-end using {\tt raw images} (light blue). We also show the results for a {\tt baseline} agent (blue) that uses simple 1-\textit{d} goal and position representations.}
\label{rl-train}
\vspace{-2mm}
\end{figure*}

\textbf{Goal-oriented navigation}: We evaluate the RL training performance on both the traversal used for training and other two testing traversals with extreme visual changes (e.g. day to afternoon/night for Oxford RobotCar, and summer to winter/fall for Nordland). We also limit the maximum number of agent steps in an episode to the number of images within the traversal, measuring in this way how well the agent can find a target destination with a moderate, environment-appropriate number of steps. We provide statistics on the number of navigation tasks that our policy can achieve by reporting the percentage of the deployment results in two categories: (1) \textit{completed} tasks, when the agent reaches the target using the minimum number of steps as defined above, or (2) \textit{failed} tasks, otherwise.

\section{Experiments: Results}

We first conduct conventional, single-frame VPR experiments using our visual observations on two stationary real-world datasets (Fig. \ref{datasets}). We then use these compact place representations to train our policy network for efficiently learning goal-driven navigation tasks using CityLearn.

\subsection{Place Recognition Experiments}\label{vpr-exps}

The trade-off of using compact visual observations for VPR is shown in Fig. \ref{auc}. We report the results of our single-frame VPR experiments, as described in Secs. \ref{feat-obs} and \ref{metrics}. AUC performance decreases as we decrease the feature dimension from 4096-\textit{d} all the way to 64-\textit{d} in both NetVLAD and ResNet-50 models. We can also observe how well these networks generalize when facing small appearance variations such as summer to fall for Nordland, see Fig. \ref{auc} (left).

For Oxford RobotCar, moderate viewpoint changes Day/sunny to overcast results in lower global performance, when compared to Nordland which does not include viewpoint changes (see Fig. \ref{auc} (right)). In contrast, for extreme appearance changes, such as summer to winter or day to night, we can observe that the global AUC performance is compromised, reducing to less than half for Nordland or even to less than a quarter for Oxford RobotCar compared to small appearance changes. It is worth noting we are performing only a single-frame matching procedure here; the results may not be as good as expected for these state-of-the-art methods since multi-frame algorithmic techniques are typically incorporated on top of those single-frame results, as previously described in Sec. \ref{vpr}.

\begin{figure*}[ht]
\subfigure{
\includegraphics[width=\columnwidth]{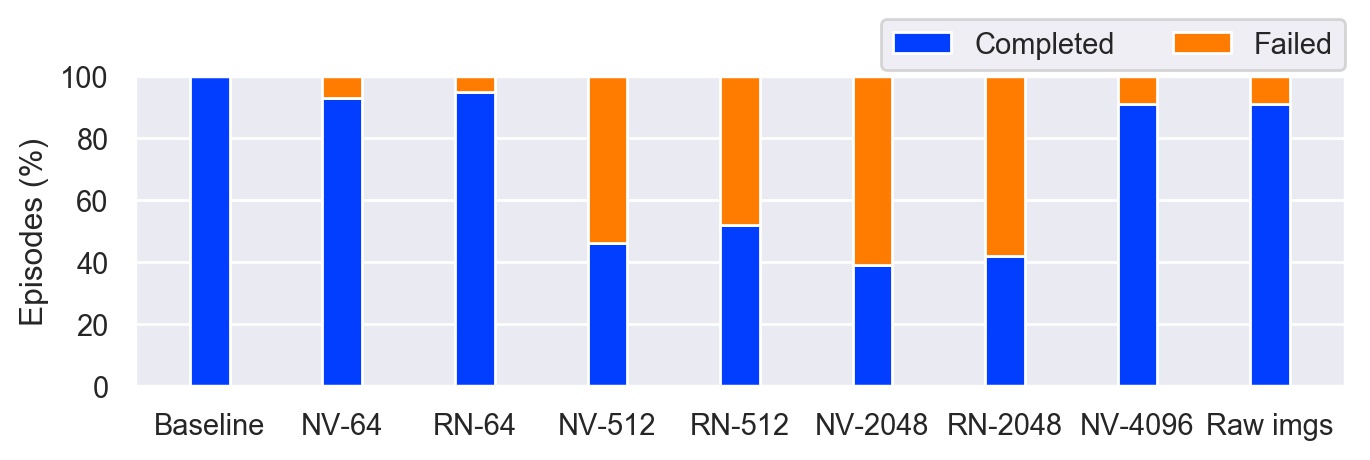}}
\subfigure{
\includegraphics[width=\columnwidth]{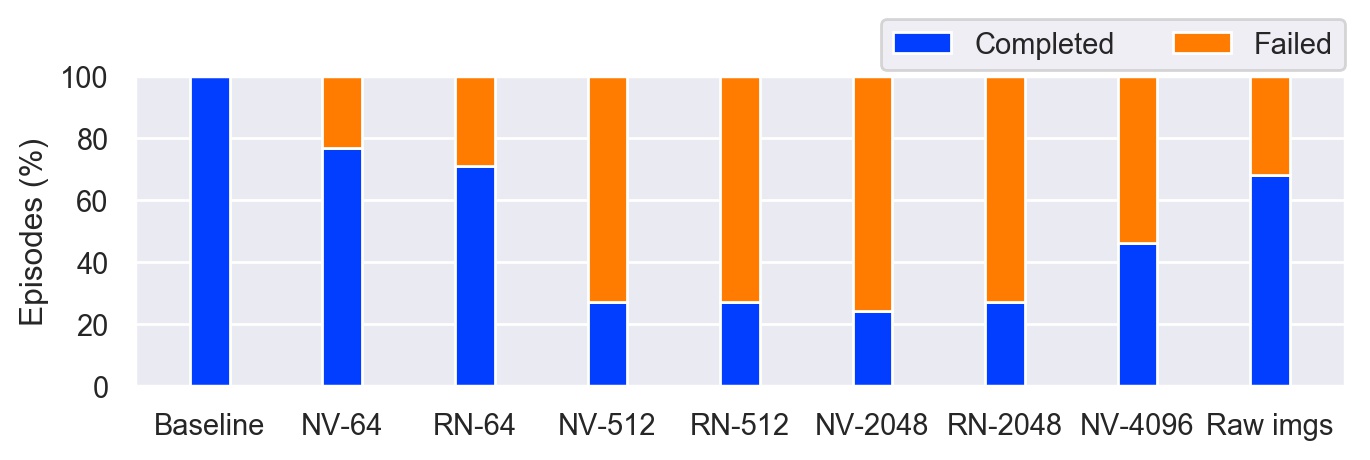}}
\vspace{-4mm}
\caption{Navigation policy evaluation statistics on the traversal it was trained. \textbf{Nordland} (left): summer, \textbf{Oxford RobotCar} (right): day.}
\label{train-eval}
\vspace{-2mm}
\end{figure*}

\begin{figure*}[ht]
\centering
\subfigure{
\includegraphics[width=\columnwidth]{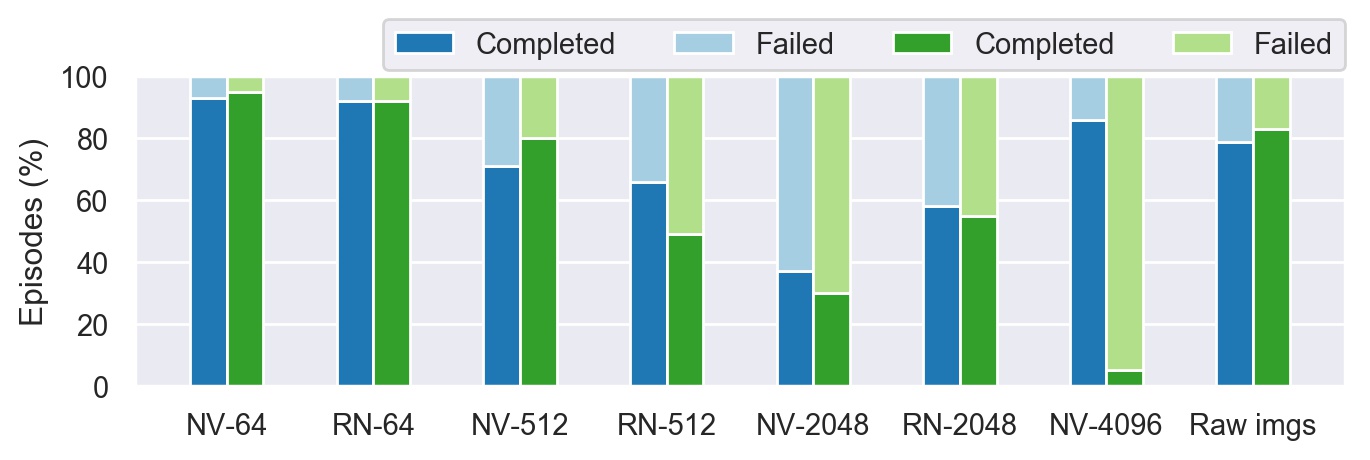}}
\subfigure{
\includegraphics[width=\columnwidth]{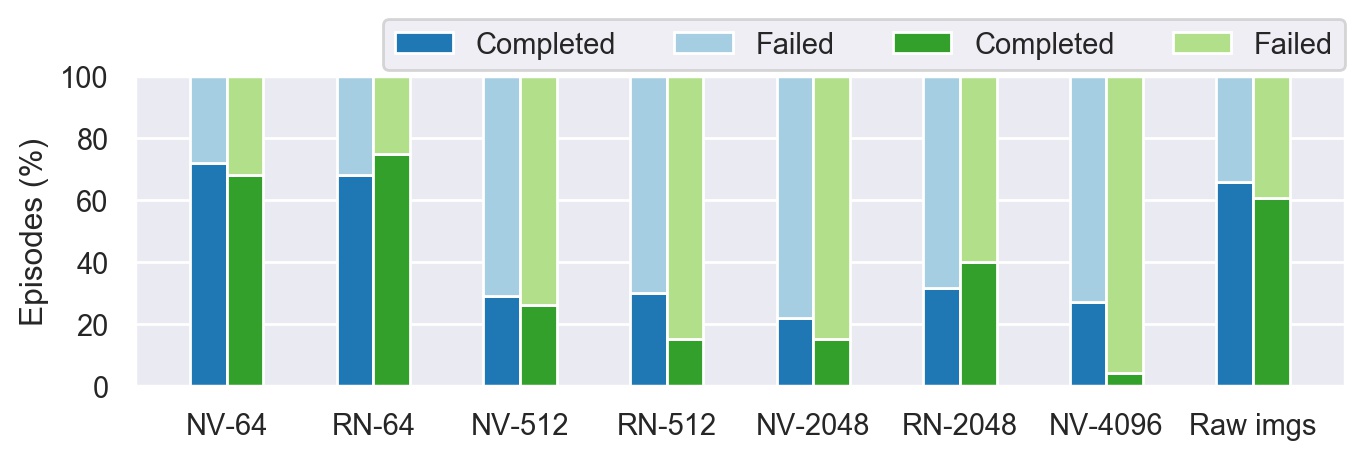}}
\vspace{-2mm}
\caption{\textbf{Generalization results}. Evaluation statistics over moderated (blue) and extreme (green) appearance environmental changes. For \textbf{Nordland} (left): fall (blue) and winter (green) traversals. For \textbf{Oxford RobotCar} (right): overcast (blue) and night (green).}
\label{test-eval}
\vspace{-2mm}
\end{figure*}

\subsection{Sample-Efficient Navigation Policy Training}

We illustrate RL training curves in Fig. \ref{rl-train}; complete-related visualization as a function of the required training time is presented in Fig. \ref{rl-tradeoff}. In Fig. \ref{rl-train} (left), we observe that {\tt our approach}, with 64-\textit{d} representations, achieves comparable average reward performance compared to the {\tt baseline} agent; being 92\% for NetVLAD, 80\% for ResNet-50 and 99\% for the {\tt baseline} agent. This small difference between these three agents is reflected in Fig. \ref{rl-train} (right), where the number of agent steps stabilizes slightly below 50 at 10,000 episodes for the {\tt baseline} agent, while for the remaining two agents (NetVLAD and ResNet-50 with 64-\textit{d}) this occurs slightly above 50 steps at 18,000 episodes.

This behavior is consequently observed again in Fig. \ref{rl-train}, as we increase the visual feature dimensions from 512-\textit{d} to 4096-\textit{d}. The final average number of steps for these agents is around 75 and the required number of training episodes increases as we increase the feature dimension; except for 4096-\textit{d} that stabilizes at 50,000 episodes, which is lower than 2048-\textit{d} that requires 60,000 episodes. We additionally provide two training results for 4096-\textit{d}, where this behavior is again shown in curve 4096-\textit{d}*. It is worth noting the training curves in Fig. \ref{rl-train} were obtained by averaging 5 trials each using different seed numbers, and then applying curve smoothing with weight 0.9 to enable cleaner visualization of our results. In all our experiments, we used 16 concurrent agents for training our policy network using the CityLearn framework.

\subsection{Deployment and Generalization}

We report evaluation statistics of our trained navigation policy on both the reference traversal used for training (Fig. \ref{train-eval}) and query traversals used to test their generalization capabilities (Fig. \ref{test-eval}) across our two datasets alongside with the CityLearn environment using the Nordland (left) and Oxford RobotCar (right) datasets. We evaluated our trained stochastic policy every 100 episodes and calculated the number of \textit{completed} and \textit{failed} navigation tasks. From Figs. \ref{train-eval} and \ref{test-eval}, it can be observed that when using compact representations (64-\textit{d}) we can achieve better generalization results, even under extreme environmental changes such as summer to winter for Nordland, shown in Fig. \ref{test-eval} (left), or day to night for Oxford RobotCar, see Fig. \ref{test-eval} (right). While increasing the feature dimension in VPR tasks results in better AUC performance (see Fig. \ref{auc}), the opposite seems to occur for navigation tasks; smaller representations are better for both final average performance and sample efficiency in terms of training time or number of episodes, as well as in generalization capabilities, at least in an RL context.

\begin{figure}[!t]
\centering
\includegraphics[width=\columnwidth]{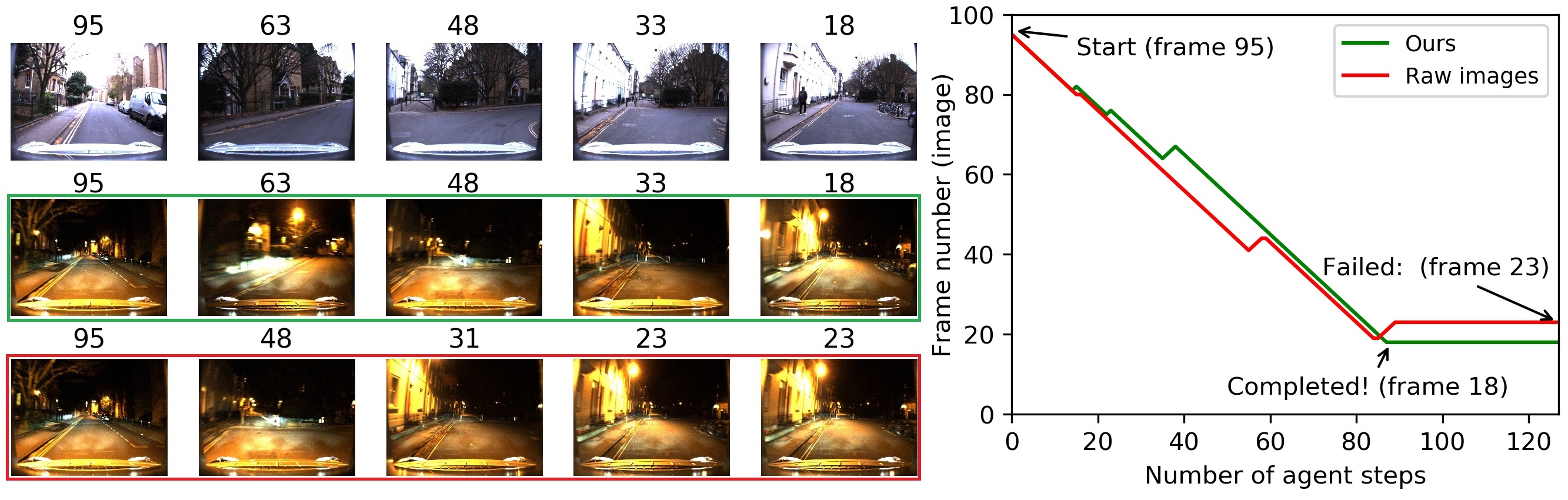}
\vspace{-6mm}
\caption{\textbf{Deployment comparison}. The policies were trained on the Oxford RobotCar dataset (day: top-left) and evaluated under extreme visual changes (night). {\tt Our approach} completed the task using NetVLAD 64-\textit{d} (center-left), while the {\tt raw images} agent failed (bottom-left).}
\label{deploy}
\vspace{-4mm}
\end{figure}

Fig. \ref{deploy} shows deployment comparisons between our approach and an agent trained end-to-end using raw images (using the policy network described in Sec. \ref{baselines}) in a route of the Oxford RobotCar dataset. Both agents starting at the same location with a common goal.

\section{Conclusions}

We conducted comprehensive experiments applying VPR and RL techniques to examine the value of using visual and self-motion (in terms of the agent's previous actions) sensory feedback to learn navigation policies on diverse robotic datasets. To enable efficient RL training, we use VPR models to encode real sensory data that, when combined with the goal destination, generates compact bimodal representations. Once trained, we showed that smaller visual representations such as 64-\textit{d} generalized better than larger features over a range of environmental transitions, while being around 2 orders of magnitude faster and requiring a small fraction of the amount of experience in terms of training time.

The proposed interactive environment, CityLearn, can also be used to load any other benchmark dataset (or even custom in-house recorded data), such as those from drones or underwater robots, to train and test many different types of navigation algorithms as well as to further build and investigate the performance of advanced RL algorithms using realistic images. Future research could include other RL algorithms such as \cite{IEEEexample:impala}, modular architectures for transfer learning to new cities \cite{IEEEexample:c15}, and adding more functionality to the environment such as creating 2D geometric maps.



\bibliographystyle{IEEEtran}
\balance

\bibliography{IEEEexample}

\begin{thebibliography}{10}
\providecommand{\url}[1]{#1}
\csname url@rmstyle\endcsname
\providecommand{\newblock}{\relax}
\providecommand{\bibinfo}[2]{#2}
\providecommand\BIBentrySTDinterwordspacing{\spaceskip=0pt\relax}
\providecommand\BIBentryALTinterwordstretchfactor{4}
\providecommand\BIBentryALTinterwordspacing{\spaceskip=\fontdimen2\font plus
\BIBentryALTinterwordstretchfactor\fontdimen3\font minus
  \fontdimen4\font\relax}
\providecommand\BIBforeignlanguage[2]{{%
\expandafter\ifx\csname l@#1\endcsname\relax
\typeout{** WARNING: IEEEtran.bst: No hyphenation pattern has been}%
\typeout{** loaded for the language `#1'. Using the pattern for}%
\typeout{** the default language instead.}%
\else
\language=\csname l@#1\endcsname
\fi
#2}}

\bibitem{IEEEexample:st-book}
S.~Thrun, W.~Burgard, and D.~Fox, \emph{Probabilistic Robotics}, 1st~ed., ser.
  Intelligent Robotics and Autonomous Agents.\hskip 1em plus 0.5em minus
  0.4em\relax The MIT Press, 2005.

\bibitem{IEEEexample:nomap}
P.~Mirowski, M.~Grimes, M.~Malinowski, K.~M. Hermann, K.~Anderson,
  D.~Teplyashin, K.~Simonyan, A.~Zisserman, R.~Hadsell, \emph{et~al.},
  ``Learning to navigate in cities without a map,'' in \emph{Advances in Neural
  Information Processing Systems}, 2018, pp. 2419--2430.

\bibitem{IEEEexample:complex}
P.~W. Mirowski, R.~Pascanu, F.~Viola, H.~Soyer, A.~Ballard, A.~Banino,
  M.~Denil, R.~Goroshin, L.~Sifre, K.~Kavukcuoglu, D.~Kumaran, and R.~Hadsell,
  ``Learning to navigate in complex environments,'' \emph{arXiv preprint
  arXiv:1611.03673}, 2016.

\bibitem{IEEEexample:banino2018vector}
A.~Banino, C.~Barry, B.~Uria, C.~Blundell, T.~Lillicrap, P.~Mirowski,
  A.~Pritzel, M.~J. Chadwick, T.~Degris, J.~Modayil, \emph{et~al.},
  ``Vector-based navigation using grid-like representations in artificial
  agents,'' \emph{Nature}, vol. 557, no. 7705, pp. 429--433, 2018.

\bibitem{IEEEexample:cueva2018emergence}
C.~J. Cueva and X.-X. Wei, ``Emergence of grid-like representations by training
  recurrent neural networks to perform spatial localization,'' \emph{arXiv
  preprint arXiv:1803.07770}, 2018.

\bibitem{IEEEexample:nvlad}
R.~Arandjelovi{\'c}, P.~Gron{\'a}t, A.~Torii, T.~Pajdla, and J.~Sivic,
  ``{NetVLAD: CNN Architecture for Weakly Supervised Place Recognition},''
  \emph{2016 IEEE Conference on Computer Vision and Pattern Recognition
  (CVPR)}, pp. 5297--5307, 2016.

\bibitem{IEEEexample:zetao2017}
Z.~Chen, A.~Jacobson, N.~S{\"u}nderhauf, B.~Upcroft, L.~Liu, C.~Shen, I.~D.
  Reid, and M.~Milford, ``Deep learning features at scale for visual place
  recognition,'' \emph{2017 IEEE International Conference on Robotics and
  Automation (ICRA)}, pp. 3223--3230, 2017.

\bibitem{IEEEexample:lost}
S.~Garg, N.~S{\"u}enderhauf, and M.~Milford, ``Lost? appearance-invariant place
  recognition for opposite viewpoints using visual semantics,''
  \emph{Proceedings of Robotics: Science and Systems XIV}, 2018.

\bibitem{IEEEexample:mpf}
S.~Hausler, A.~Jacobson, and M.~Milford, ``Multi-process fusion: Visual place
  recognition using multiple image processing methods,'' \emph{IEEE Robotics
  and Automation Letters}, vol.~4, no.~2, pp. 1924--1931, 2019.

\bibitem{IEEEexample:nav101}
H.~F. Durrant-Whyte, ``An autonomous guided vehicle for cargo handling
  applications,'' \emph{The International Journal of Robotics Research},
  vol.~15, pp. 407--440, 1996.

\bibitem{IEEEexample:nav102}
I.~R. Nourbakhsh, J.~Bobenage, S.~Grange, R.~Lutz, R.~Meyer, and A.~Soto, ``An
  affective mobile robot educator with a full-time job,'' \emph{Artificial
  Intelligence}, vol. 114, no.~1, pp. 95--124, 1999.

\bibitem{IEEEexample:st98}
F.~{Dellaert}, D.~{Fox}, W.~{Burgard}, and S.~{Thrun}, ``Monte carlo
  localization for mobile robots,'' in \emph{Proceedings 1999 IEEE
  International Conference on Robotics and Automation (ICRA)}, vol.~2, May
  1999, pp. 1322--1328.

\bibitem{IEEEexample:lin}
{L. Lin}, ``Reinforcement learning for robots using neural networks,'' Ph.D.
  dissertation, Carnegie Mellon University, Pittsburgh, PA, 1992.

\bibitem{IEEEexample:c7}
\BIBentryALTinterwordspacing
Y.~Bengio, J.~Louradour, R.~Collobert, and J.~Weston, ``Curriculum learning,''
  in \emph{Proceedings of the 26th Annual International Conference on Machine
  Learning}, ser. ICML ’09.\hskip 1em plus 0.5em minus 0.4em\relax New York,
  NY, USA: Association for Computing Machinery, 2009, pp. 41--48. [Online].
  Available: \url{https://doi.org/10.1145/1553374.1553380}
\BIBentrySTDinterwordspacing

\bibitem{IEEEexample:cnn1}
I.~G. Goodfellow, Y.~Bengio, and A.~C. Courville, ``Deep learning,''
  \emph{Nature}, vol. 521, pp. 436--444, 2015.

\bibitem{IEEEexample:deepnav}
S.~{Brahmbhatt} and J.~{Hays}, ``Deepnav: Learning to navigate large cities,''
  in \emph{2017 IEEE Conference on Computer Vision and Pattern Recognition
  (CVPR)}, July 2017, pp. 3087--3096.

\bibitem{IEEEexample:c20}
A.~Khosla, B.~An~An, J.~J. Lim, and A.~Torralba, ``Looking beyond the visible
  scene,'' in \emph{Proceedings of the IEEE Conference on Computer Vision and
  Pattern Recognition}, 2014, pp. 3710--3717.

\bibitem{IEEEexample:carballo2018end}
A.~Carballo, S.~Seiya, J.~Lambert, H.~Darweesh, P.~Narksri, L.~Y. Morales,
  N.~Akai, E.~Takeuchi, and K.~Takeda, ``End-to-end autonomous mobile robot
  navigation with model-based system support,'' \emph{Journal of Robotics and
  Mechatronics}, vol.~30, no.~4, pp. 563--583, 2018.

\bibitem{IEEEexample:amini2019variational}
A.~Amini, G.~Rosman, S.~Karaman, and D.~Rus, ``Variational end-to-end
  navigation and localization,'' in \emph{2019 International Conference on
  Robotics and Automation (ICRA)}, 2019, pp. 8958--8964.

\bibitem{IEEEexample:xu2017end}
H.~Xu, Y.~Gao, F.~Yu, and T.~Darrell, ``End-to-end learning of driving models
  from large-scale video datasets,'' in \emph{Proceedings of the IEEE
  Conference on Computer Vision and Pattern Recognition (CVPR)}, 2017, pp.
  2174--2182.

\bibitem{IEEEexample:c12}
H.~de~Vries, K.~Shuster, D.~Batra, D.~Parikh, J.~Weston, and D.~Kiela, ``Talk
  the walk: Navigating new york city through grounded dialogue,'' \emph{arXiv
  preprint arXiv:1807.03367}, 2018.

\bibitem{IEEEexample:c13}
V.~Cirik, Y.~Zhang, and J.~Baldridge, ``Following formulaic map instructions in
  a street simulation environment,'' in \emph{2018 NeurIPS Workshop on Visually
  Grounded Interaction and Language}, vol.~1, 2018.

\bibitem{IEEEexample:c14}
H.~Chen, A.~Suhr, D.~Misra, N.~Snavely, and Y.~Artzi, ``Touchdown: Natural
  language navigation and spatial reasoning in visual street environments,'' in
  \emph{Proceedings of the IEEE Conference on Computer Vision and Pattern
  Recognition}, 2019, pp. 12\,538--12\,547.

\bibitem{IEEEexample:dhiman2018critical}
V.~Dhiman, S.~Banerjee, B.~Griffin, J.~M. Siskind, and J.~J. Corso, ``A
  critical investigation of deep reinforcement learning for navigation,''
  \emph{arXiv preprint arXiv:1802.02274}, 2018.

\bibitem{IEEEexample:deplo}
\BIBentryALTinterwordspacing
J.~Bruce, N.~S{\"u}nderhauf, P.~Mirowski, R.~Hadsell, and M.~Milford,
  ``Learning deployable navigation policies at kilometer scale from a single
  traversal,'' in \emph{Proceedings of The 2nd Conference on Robot Learning},
  ser. Proceedings of Machine Learning Research, A.~Billard, A.~Dragan,
  J.~Peters, and J.~Morimoto, Eds., vol.~87.\hskip 1em plus 0.5em minus
  0.4em\relax PMLR, 29--31 Oct 2018, pp. 346--361. [Online]. Available:
  \url{http://proceedings.mlr.press/v87/bruce18a.html}
\BIBentrySTDinterwordspacing

\bibitem{IEEEexample:c16}
Y.~Zhu, R.~Mottaghi, E.~Kolve, J.~J. Lim, A.~Gupta, L.~Fei-Fei, and A.~Farhadi,
  ``Target-driven visual navigation in indoor scenes using deep reinforcement
  learning,'' in \emph{2017 IEEE international conference on robotics and
  automation (ICRA)}, 2017, pp. 3357--3364.

\bibitem{IEEEexample:gupta2017cognitive}
S.~Gupta, J.~Davidson, S.~Levine, R.~Sukthankar, and J.~Malik, ``Cognitive
  mapping and planning for visual navigation,'' in \emph{Proceedings of the
  IEEE Conference on Computer Vision and Pattern Recognition}, 2017, pp.
  2616--2625.

\bibitem{IEEEexample:Kolve2017AI2THORAI}
E.~Kolve, R.~Mottaghi, D.~Gordon, Y.~Zhu, A.~Gupta, and A.~Farhadi,
  ``{AI2-THOR: An Interactive 3D Environment for Visual AI},'' \emph{arXiv
  preprint arXiv:1712.05474}, 2017.

\bibitem{IEEEexample:savva2017minos}
M.~Savva, A.~X. Chang, A.~Dosovitskiy, T.~Funkhouser, and V.~Koltun, ``{MINOS:
  Multimodal indoor simulator for navigation in complex environments},''
  \emph{arXiv preprint arXiv:1712.03931}, 2017.

\bibitem{IEEEexample:kahn2018composable}
G.~Kahn, A.~Villaflor, P.~Abbeel, and S.~Levine, ``Composable
  action-conditioned predictors: Flexible off-policy learning for robot
  navigation,'' \emph{arXiv preprint arXiv:1810.07167}, 2018.

\bibitem{IEEEexample:brunner2018teaching}
G.~Brunner, O.~Richter, Y.~Wang, and R.~Wattenhofer, ``Teaching a machine to
  read maps with deep reinforcement learning,'' in \emph{Thirty-Second AAAI
  Conference on Artificial Intelligence}, 2018.

\bibitem{IEEEexample:wu2018building}
Y.~Wu, Y.~Wu, G.~Gkioxari, and Y.~Tian, ``{Building generalizable agents with a
  realistic and rich 3D environment},'' \emph{arXiv preprint arXiv:1801.02209},
  2018.

\bibitem{IEEEexample:bansal2019combining}
S.~Bansal, V.~Tolani, S.~Gupta, J.~Malik, and C.~Tomlin, ``Combining optimal
  control and learning for visual navigation in novel environments,''
  \emph{arXiv preprint arXiv:1903.02531}, 2019.

\bibitem{IEEEexample:wang2018omnidirectional}
T.-H. Wang, H.-J. Huang, J.-T. Lin, C.-W. Hu, K.-H. Zeng, and M.~Sun,
  ``Omnidirectional cnn for visual place recognition and navigation,'' in
  \emph{2018 IEEE International Conference on Robotics and Automation (ICRA)},
  2018, pp. 2341--2348.

\bibitem{IEEEexample:mousavian2019visual}
A.~Mousavian, A.~Toshev, M.~Fi{\v{s}}er, J.~Ko{\v{s}}eck{\'a}, A.~Wahid, and
  J.~Davidson, ``Visual representations for semantic target driven
  navigation,'' in \emph{2019 International Conference on Robotics and
  Automation (ICRA)}.\hskip 1em plus 0.5em minus 0.4em\relax IEEE, 2019, pp.
  8846--8852.

\bibitem{IEEEexample:c31}
D.~S. Chaplot, E.~Parisotto, and R.~Salakhutdinov, ``Active neural
  localization,'' in \emph{International Conference on Learning
  Representations}, 2018.

\bibitem{IEEEexample:seqslam}
M.~J. {Milford} and G.~F. {Wyeth}, ``{SeqSLAM: Visual route-based navigation
  for sunny summer days and stormy winter nights},'' in \emph{2012 IEEE
  International Conference on Robotics and Automation (ICRA)}, May 2012, pp.
  1643--1649.

\bibitem{IEEEexample:chen2014convolutional}
Z.~Chen, O.~Lam, A.~Jacobson, and M.~Milford, ``Convolutional neural
  network-based place recognition,'' \emph{arXiv preprint arXiv:1411.1509},
  2014.

\bibitem{IEEEexample:naseer2014robust}
T.~Naseer, L.~Spinello, W.~Burgard, and C.~Stachniss, ``Robust visual robot
  localization across seasons using network flows,'' in \emph{Twenty-eighth
  AAAI conference on artificial intelligence}, 2014.

\bibitem{IEEEexample:pepperell2014all}
E.~Pepperell, P.~I. Corke, and M.~J. Milford, ``All-environment visual place
  recognition with smart,'' in \emph{IEEE International Conference on Robotics
  and Automation (ICRA)}, 2014, pp. 1612--1618.

\bibitem{IEEEexample:milford2014condition}
M.~Milford, W.~Scheirer, E.~Vig, A.~Glover, O.~Baumann, J.~Mattingley, and
  D.~Cox, ``Condition-invariant, top-down visual place recognition,'' in
  \emph{2014 IEEE International Conference on Robotics and Automation (ICRA)},
  2014, pp. 5571--5577.

\bibitem{IEEEexample:vprsurvey}
S.~{Lowry}, N.~{Sünderhauf}, P.~{Newman}, J.~J. {Leonard}, D.~{Cox},
  P.~{Corke}, and M.~J. {Milford}, ``Visual place recognition: A survey,''
  \emph{IEEE Transactions on Robotics}, vol.~32, no.~1, pp. 1--19, Feb 2016.

\bibitem{IEEEexample:sunderhauf2015performance}
N.~S{\"u}nderhauf, S.~Shirazi, F.~Dayoub, B.~Upcroft, and M.~Milford, ``On the
  performance of convnet features for place recognition,'' in \emph{2015
  IEEE/RSJ International Conference on Intelligent Robots and Systems (IROS)},
  2015, pp. 4297--4304.

\bibitem{IEEEexample:cnnlanmark}
N.~S{\"u}nderhauf, S.~Shirazi, A.~Jacobson, F.~Dayoub, E.~Pepperell,
  B.~Upcroft, and M.~Milford, ``Place recognition with convnet landmarks:
  Viewpoint-robust, condition-robust, training-free,'' \emph{Proceedings of
  Robotics: Science and Systems XII}, 2015.

\bibitem{IEEEexample:ferrarini2019visual}
B.~Ferrarini, M.~Waheed, S.~Waheed, S.~Ehsan, M.~Milford, and K.~D.
  McDonald-Maier, ``Visual place recognition for aerial robotics: Exploring
  accuracy-computation trade-off for local image descriptors,'' in \emph{2019
  NASA/ESA Conference on Adaptive Hardware and Systems (AHS)}, 2019, pp.
  103--108.

\bibitem{IEEEexample:merrill2018lightweight}
N.~Merrill and G.~Huang, ``Lightweight unsupervised deep loop closure,''
  \emph{arXiv preprint arXiv:1805.07703}, 2018.

\bibitem{IEEEexample:lowry2018lightweight}
S.~Lowry and H.~Andreasson, ``Lightweight, viewpoint-invariant visual place
  recognition in changing environments,'' \emph{IEEE Robotics and Automation
  Letters}, vol.~3, no.~2, pp. 957--964, 2018.

\bibitem{IEEEexample:sattler2018benchmarking}
T.~Sattler, W.~Maddern, C.~Toft, A.~Torii, L.~Hammarstrand, E.~Stenborg,
  D.~Safari, M.~Okutomi, M.~Pollefeys, J.~Sivic, \emph{et~al.}, ``Benchmarking
  6dof outdoor visual localization in changing conditions,'' in
  \emph{Proceedings of the IEEE Conference on Computer Vision and Pattern
  Recognition}, 2018, pp. 8601--8610.

\bibitem{IEEEexample:mount2019automatic}
J.~Mount, L.~Dawes, and M.~J. Milford, ``Automatic coverage selection for
  surface-based visual localization,'' \emph{IEEE Robotics and Automation
  Letters}, vol.~4, no.~4, pp. 3900--3907, 2019.

\bibitem{IEEEexample:arelarge}
\BIBentryALTinterwordspacing
A.~{Torii}, H.~{Taira}, J.~{Sivic}, M.~{Pollefeys}, M.~{Okutomi}, T.~{Pajdla},
  and T.~{Sattler}, ``{Are Large-Scale 3D Models Really Necessary for Accurate
  Visual Localization?}'' \emph{IEEE Transactions on Pattern Analysis and
  Machine Intelligence}, 2019. [Online]. Available:
  \url{https://doi.org/10.1109/TPAMI.2019.2941876}
\BIBentrySTDinterwordspacing

\bibitem{IEEEexample:hausler2019filter}
S.~Hausler, A.~Jacobson, and M.~Milford, ``Filter early, match late: Improving
  network-based visual place recognition,'' \emph{arXiv preprint
  arXiv:1906.12176}, 2019.

\bibitem{IEEEexample:lowry2019similarity}
S.~Lowry, ``Similarity criteria: evaluating perceptual change for visual
  localization,'' in \emph{2019 European Conference on Mobile Robots (ECMR)},
  2019, pp. 1--6.

\bibitem{IEEEexample:khaliq2019holistic}
\BIBentryALTinterwordspacing
A.~{Khaliq}, S.~{Ehsan}, Z.~{Chen}, M.~{Milford}, and K.~{McDonald-Maier}, ``A
  holistic visual place recognition approach using lightweight cnns for
  significant viewpoint and appearance changes,'' \emph{IEEE Transactions on
  Robotics}, pp. 1--9, 2019. [Online]. Available:
  \url{https://doi.org/10.1109/TRO.2019.2956352}
\BIBentrySTDinterwordspacing

\bibitem{IEEEexample:chancan2020hybrid}
M.~{Chanc\'an}, L.~{Hernandez-Nunez}, A.~{Narendra}, A.~B. {Barron}, and
  M.~{Milford}, ``A hybrid compact neural architecture for visual place
  recognition,'' \emph{IEEE Robotics and Automation Letters}, vol.~5, no.~2,
  pp. 993--1000, April 2020.

\bibitem{IEEEexample:openseqslam2}
B.~{Talbot}, S.~{Garg}, and M.~{Milford}, ``{OpenSeqSLAM2.0: An Open Source
  Toolbox for Visual Place Recognition Under Changing Conditions},'' in
  \emph{2018 IEEE/RSJ International Conference on Intelligent Robots and
  Systems (IROS)}, Oct 2018, pp. 7758--7765.

\bibitem{IEEEexample:garg2018don}
S.~Garg, N.~Suenderhauf, and M.~Milford, ``Don't look back: Robustifying place
  categorization for viewpoint-and condition-invariant place recognition,'' in
  \emph{2018 IEEE International Conference on Robotics and Automation (ICRA)},
  2018, pp. 3645--3652.

\bibitem{IEEEexample:resnet}
K.~{He}, X.~{Zhang}, S.~{Ren}, and J.~{Sun}, ``Deep residual learning for image
  recognition,'' in \emph{2016 IEEE Conference on Computer Vision and Pattern
  Recognition (CVPR)}, June 2016, pp. 770--778.

\bibitem{IEEEexample:levelling}
M.~Zaffar, A.~Khaliq, S.~Ehsan, M.~Milford, and K.~D. McDonald-Maier,
  ``Levelling the playing field: A comprehensive comparison of visual place
  recognition approaches under changing conditions,'' \emph{arXiv preprint
  arXiv:1903.09107}, 2019.

\bibitem{IEEEexample:mirowski2019streetlearn}
P.~Mirowski, A.~Banki-Horvath, K.~Anderson, D.~Teplyashin, K.~M. Hermann,
  M.~Malinowski, M.~K. Grimes, K.~Simonyan, K.~Kavukcuoglu, A.~Zisserman,
  \emph{et~al.}, ``The streetlearn environment and dataset,'' \emph{arXiv
  preprint arXiv:1903.01292}, 2019.

\bibitem{IEEEexample:oxford}
\BIBentryALTinterwordspacing
W.~Maddern, G.~Pascoe, C.~Linegar, and P.~Newman, ``1 year, 1000 km: The oxford
  robotcar dataset,'' \emph{The International Journal of Robotics Research},
  vol.~36, no.~1, pp. 3--15, 2017. [Online]. Available:
  \url{https://doi.org/10.1177/0278364916679498}
\BIBentrySTDinterwordspacing

\bibitem{IEEEexample:lookonce}
Z.~Chen, F.~Maffra, I.~Sa, and M.~Chli, ``Only look once, mining distinctive
  landmarks from convnet for visual place recognition,'' in \emph{2017 IEEE/RSJ
  International Conference on Intelligent Robots and Systems (IROS)}, 2017, pp.
  9--16.

\bibitem{IEEEexample:fabrat}
A.~J. {Glover}, W.~P. {Maddern}, M.~J. {Milford}, and G.~F. {Wyeth}, ``{FAB-MAP
  + RatSLAM: Appearance-based SLAM for multiple times of day},'' in \emph{2010
  IEEE International Conference on Robotics and Automation}, May 2010, pp.
  3507--3512.

\bibitem{IEEEexample:seqsearch}
M.~{Milford}, S.~{Lowry}, N.~{Sunderhauf}, S.~{Shirazi}, E.~{Pepperell},
  B.~{Upcroft}, C.~{Shen}, G.~{Lin}, F.~{Liu}, C.~{Cadena}, and I.~{Reid},
  ``Sequence searching with deep-learnt depth for condition- and
  viewpoint-invariant route-based place recognition,'' in \emph{2015 IEEE
  Conference on Computer Vision and Pattern Recognition Workshops (CVPRW)},
  June 2015, pp. 18--25.

\bibitem{IEEEexample:kitti}
\BIBentryALTinterwordspacing
A.~Geiger, P.~Lenz, C.~Stiller, and R.~Urtasun, ``{Vision meets robotics: The
  KITTI dataset},'' \emph{The International Journal of Robotics Research},
  vol.~32, no.~11, pp. 1231--1237, 2013. [Online]. Available:
  \url{https://doi.org/10.1177/0278364913491297}
\BIBentrySTDinterwordspacing

\bibitem{IEEEexample:naseer}
T.~{Naseer}, W.~{Burgard}, and C.~{Stachniss}, ``Robust visual localization
  across seasons,'' \emph{IEEE Transactions on Robotics}, vol.~34, no.~2, pp.
  289--302, April 2018.

\bibitem{IEEEexample:guo2018safe}
J.~Guo, U.~Kurup, and M.~Shah, ``Is it safe to drive? an overview of factors,
  challenges, and datasets for driveability assessment in autonomous driving,''
  \emph{arXiv preprint arXiv:1811.11277}, 2018.

\bibitem{IEEEexample:caesar2019nuscenes}
H.~Caesar, V.~Bankiti, A.~H. Lang, S.~Vora, V.~E. Liong, Q.~Xu, A.~Krishnan,
  Y.~Pan, G.~Baldan, and O.~Beijbom, ``{nuScenes: A multimodal dataset for
  autonomous driving},'' \emph{arXiv preprint arXiv:1903.11027}, 2019.

\bibitem{IEEEexample:juliani2018unity}
\BIBentryALTinterwordspacing
A.~Juliani, V.-P. Berges, E.~Vckay, Y.~Gao, H.~Henry, M.~Mattar, and D.~Lange,
  ``Unity: A general platform for intelligent agents,'' \emph{arXiv preprint
  arXiv:1809.02627}, 2018. [Online]. Available:
  \url{https://github.com/Unity-Technologies/ml-agents}
\BIBentrySTDinterwordspacing

\bibitem{IEEEexample:hermann2019learning}
K.~M. Hermann, M.~Malinowski, P.~Mirowski, A.~Banki-Horvath, K.~Anderson, and
  R.~Hadsell, ``Learning to follow directions in street view,'' \emph{arXiv
  preprint arXiv:1903.00401}, 2019.

\bibitem{IEEEexample:li2019cross}
A.~Li, H.~Hu, P.~Mirowski, and M.~Farajtabar, ``Cross-view policy learning for
  street navigation,'' in \emph{Proceedings of the IEEE International
  Conference on Computer Vision}, 2019, pp. 8100--8109.

\bibitem{IEEEexample:nordland}
N.~S{\"u}nderhauf, P.~Neubert, and P.~Protzel, ``Are we there yet? challenging
  seqslam on a 3000 km journey across all four seasons,'' in \emph{Proc.
  Workshop Long-Term Autonomy 2013 IEEE Int. Conf. Robot. Autom. (ICRA)}, 2013.

\bibitem{IEEEexample:simonyan2014very}
K.~Simonyan and A.~Zisserman, ``Very deep convolutional networks for
  large-scale image recognition,'' \emph{arXiv preprint arXiv:1409.1556}, 2014.

\bibitem{IEEEexample:imagenet}
J.~{Deng}, W.~{Dong}, R.~{Socher}, L.~{Li}, {Kai Li}, and {Li Fei-Fei},
  ``Imagenet: A large-scale hierarchical image database,'' in \emph{2009 IEEE
  Conference on Computer Vision and Pattern Recognition}, June 2009, pp.
  248--255.

\bibitem{IEEEexample:ppo}
J.~Schulman, F.~Wolski, P.~Dhariwal, A.~Radford, and O.~Klimov, ``Proximal
  policy optimization algorithms,'' \emph{arXiv preprint arXiv:1707.06347},
  2017.

\bibitem{IEEEexample:trpo}
J.~Schulman, S.~Levine, P.~Abbeel, M.~Jordan, and P.~Moritz, ``Trust region
  policy optimization,'' in \emph{International Conference on Machine
  Learning}, 2015, pp. 1889--1897.

\bibitem{IEEEexample:lstm}
S.~Hochreiter and J.~Schmidhuber, ``Long short-term memory,'' \emph{Neural
  Computation}, vol.~9, pp. 1735--1780, 1997.

\bibitem{IEEEexample:impala}
L.~Espeholt, H.~Soyer, R.~Munos, K.~Simonyan, V.~Mnih, T.~Ward, Y.~Doron,
  V.~Firoiu, T.~Harley, I.~Dunning, S.~Legg, and K.~Kavukcuoglu, ``{IMPALA:
  Scalable Distributed Deep-RL with Importance Weighted Actor-Learner
  Architectures},'' \emph{arXiv preprint arXiv:1802.01561}, 2018.

\bibitem{IEEEexample:c15}
C.~Devin, A.~Gupta, T.~Darrell, P.~Abbeel, and S.~Levine, ``Learning modular
  neural network policies for multi-task and multi-robot transfer,'' in
  \emph{2017 IEEE International Conference on Robotics and Automation (ICRA)},
  2017, pp. 2169--2176.

\end{thebibliography}
\addtolength{\textheight}{-5cm}


\end{document}